\documentclass[11pt,a4paper]{article}
\usepackage[hyperref]{emnlp2020}
\usepackage[utf8]{inputenc} 
\usepackage[LGR,T1]{fontenc} 
\usepackage{times}
\usepackage{latexsym}
\usepackage{tikz}
\usepackage{tikzscale}
\usepackage{pdflscape}
\usepackage{amsmath}
\usepackage{gensymb} 
\usepackage{booktabs} 

\newcommand{\Sref}[1]{Section~\ref{#1}}
\newcommand{\Fref}[1]{Figure~\ref{#1}}
\newcommand{\Tref}[1]{Table~\ref{#1}}

\usepackage{microtype}

\aclfinalcopy 
\def\aclpaperid{SIGTYP 7} 

\title{Predicting Typological Features in WALS using Language Embeddings and Conditional Probabilities:
ÚFAL Submission to the SIGTYP 2020 Shared Task}

\author{Martin Vastl
  \And
  Daniel Zeman \\
  Charles University, Prague \\
  Faculty of Mathematics and Physics \\
  Institute of Formal and Applied Linguistics (ÚFAL) \\
  \texttt{martin.vastl2@gmail.com, \{zeman|rosa\}@ufal.mff.cuni.cz} \\
  \And
  Rudolf Rosa}

\date{}

\begin{document}
\maketitle
\begin{abstract}
We present our submission to the SIGTYP 2020 Shared Task on the prediction
of typological features. We submit a constrained system, predicting
typological features only based on the WALS database. We investigate two
approaches. The simpler of the two is a system based on estimating
correlation of feature values within languages by computing conditional
probabilities and mutual information. The second approach is to train a
neural predictor operating on precomputed language embeddings based on WALS
features. Our submitted system combines the two approaches based on their
self-estimated confidence scores. We reach the accuracy of 70.7\% on the
test data and rank first in the shared task.
\end{abstract}

\section{Introduction}

The World Atlas of Language Structures (WALS) \cite{wals} is a database of
over 2,000 languages, which lists structural properties (`features') of each
language, gathered from reference grammars. The properties/features are
phonological (such as the number of distinct consonants), grammatical (such
as morphological devices or dominant word order) and lexical (such as the
inventory of lexemes for colors). WALS can be browsed
online\footnote{\url{https://wals.info/}} and the database is also available
for download. It has been used in cross-lingual NLP to identify languages
that are grammatically similar, and to cope with expected dissimilarities
\cite{ohoran2016}.

Unfortunately the WALS database is sparse as many feature values for many
languages are missing. The goal of the present shared task is to predict the
missing features with the help of the information that is known. Some
typological properties are mutually dependent \cite{greenberg} and
implications among the feature values can be found \cite{daume2007}.

We try to learn these implications from the WALS data using a combination of
two machine learning approaches. After summarizing the task setup in
\Sref{sec:task}, we describe our systems in \Sref{sec:systems}, including
models which we tested but did not use in the final submission. We thoroughly
evaluate the systems in \Sref{sec:eval}.

\section{Task and Data}
\label{sec:task}

The SIGTYP 2020 shared task \cite{st-overview2020sigtyp} splits the WALS data
into training, development and test portion. In the blind test data, some
feature values are masked and the participating system is supposed to predict
them based on the remaining features that are left visible.

The task specification envisions a constrained and an unconstrained track,
where the constrained systems can use only the provided WALS data, while an
unconstrained system can use additional external resources, such as texts or
pre-trained word vectors. We stay within the constrained track as for most
languages we do not have any other data anyway.

The training data consists of 1,125 languages, the development data has 83
and the test data has 149 languages. Each language has 7 \textit{general
properties}, which are always filled in: the WALS language code, language
name, genus, family, latitude, longitude and country codes (more than one
country can be listed for a language). Then there are up to 185
\textit{linguistic properties}. No language has all of them and no linguistic
property has a known value for all languages. Within the training data, the
best described language is English with 159 linguistic properties; the least
described languages have just four linguistic properties each, and the median
is 28. The most covered linguistic property is 83A \textit{Order of Object
and Verb}, filled in for 785 languages; the median is 168 languages, and the
two least covered properties appear with 8 languages each. We refer to all
language properties (general and linguistic) as \textit{features}.

To be able to use the development data for evaluation, we randomly masked
about one half of the linguistic properties (equivalent to about 42\% of all
features) and let the system predict them. On average, a language had 26.9
non-empty features and 19.2 features to predict. Minimum knowledge was 8
non-empty features (i.e., the 7 general properties plus one linguistic
property), with 3 features to predict. Note that the organizers did not
reveal in advance what would be the proportion of known and unknown features
in the blind test data; in the end it turned out to be very similar to the
proportion we used in our development data.

\section{Systems}
\label{sec:systems}

We tried several approaches to the problem, two of which yielded promising
results. We refer to them as the Probabilistic System
(\Sref{sec:probabilistic}) and the Neural System (\Sref{sec:neural}). Our
final setup is a combination of the two (\Sref{sec:combination}); the output
of this setup was submitted to the shared task evaluation. We also briefly
mention other setups that we considered and abandoned because they did not
fare well in preliminary evaluation (\Sref{sec:other}).

\subsection{The Probabilistic System}
\label{sec:probabilistic}

Our simplest system works directly with the assumption that there are
correlations between individual language properties. This assumption is
widely accepted in typology, instantiated in particular in the Greenberg
universals \cite{greenberg}. For example, universal number 17 says that
``with overwhelmingly more than chance frequency, languages with dominant
order VSO have the adjective after the noun.'' Carried over to WALS, the
universal implies that if feature 81A \textit{Order of Subject, Object and
Verb} has value \textit{3 VSO}, then feature 87A \textit{Order of Adjective
and Noun} should have the value \textit{2 Noun-Adjective}. If we know the
universal and we see that the dominant word order in a language is VSO while
the value of 87A is unknown, we can guess its value with high confidence.
However, we do not let our model look at the list of Greenberg universals.
Instead, we try to learn similar implications directly from the WALS training
data. Specifically for the universal 17, the training data confirms the
tendency claimed by Greenberg, although it definitely does not guarantee
100\% prediction accuracy: out of 54 training languages for which both
features are known and their dominant order is VSO, 28 languages (52\%) have
adjectives after nouns, 19 languages (35\%) have adjectives before nouns, and
7 languages (13\%) have no dominant adjective-noun order.

\subsubsection{Model}
\label{sec:probabilistic:model}

Formally speaking, we have language $L$ and features $s$ and $t$. The value
of feature $s$ for language $L$ is known: $s(L)=x$. The value of feature $t$
is unknown for this language: $t(L)=?$. We know the value range of $t$ and we
can estimate the conditional probability of each value based on training
languages that have known values of both $s$ and $t$: $P(t=y|s=x)$.

For each unknown feature $t_j$ of a test language $L$ we look at all features
$s_i$ whose values are known for $L$. Different source features $s_i$ will
provide different predictions of $t_j$, so we compare their probabilities.
However, some probabilities are less reliable than others because they are
based on fewer observations. To counterbalance that, we score each prediction
by the multiple of its probability and the logarithm of the observation count
$c$. Correlations based on a single observation are thus ignored completely.

\begin{align}
  score_1&(s_i=x,t_j=y)\notag\\
    & = P(t_j=y|s_i=x)\notag\\
    & \times \log c(s_i=x,t_j=y)
\end{align}

\noindent Another danger lies in features that appear frequently but
contribute little information. For example, feature 143G \textit{Minor
morphological means of signaling negation} is with 750 occurrences almost as
frequent as 81A, but its values are very unbalanced: for 99\% languages the
value is \textit{4 None}. Co-occurrences of this value with 87A may seem to
provide a strong signal that a given language prefers adjectives after nouns,
but in reality they only reflect the general observation that there are more
languages in WALS with post-nominal adjectives than the opposite. Therefore
we compute the mutual information $I$ of the probability distributions of
each pair of features, and we add it as a third scoring factor:

\begin{align}
  score_2&(s_i=x,t_j=y)\notag\\
    & = P(t_j=y|s_i=x)\notag\\
    & \times \log c(s_i=x,t_j=y)\notag\\
    & \times I(s_i,t_j)
\end{align}

where

\begin{align}
  I(s_i,t_j) = \sum_{x} \sum_{y} & P(s_i=x, t_j=y)\notag\\
                                 & \times \log \frac{P(s_i=x, t_j=y)}{P(s_i=x)P(t_j=y)}
\end{align}

\noindent For example, $I(\text{143G},\text{87A})=0.004$ while
$I(\text{81A},\text{87A})=0.072$, meaning that the order of subject, object
and verb is a much better indicator for the order of adjective and noun than
the minor morphological means of negation.

Experiments with the development data have shown that $score_2$ (with mutual
information) gives better results than $score_1$ (without mutual
information), and both are better than the conditional probability alone. In
all three settings we always take the source feature which provides the
best-scored prediction, and ignore all other known features of the language.
We have experimented with various voting schemes to combine signals from
multiple source features but we were not able to obtain better results than
with the single best prediction.

\subsubsection{Feature Manipulation}

Everything we know about the language can be used as a source feature.
Besides numbered linguistic properties this also includes the language
family, genus, geographical coordinates and country codes. We only do not
trust the country code `US' (United States) as we observed that it occurs
mistakenly with many languages from various parts of the world. We treat
these languages as if their country code was unknown.

The geographical coordinates also deserve special attention. They represent a
point on the map where the language is displayed in WALS. Presumably the
point is distinct from other languages in WALS and it lies near the center of
the area where speakers of the language live. There is no indication of how
large the area is; nevertheless, it is still possible that the points of
small languages are close enough to indicate potential language contact.

Since the values of latitude and longitude are rarely shared among languages,
we manually partitioned the globe to a number of unevenly-sized zones and
replaced the exact coordinate with the zone. In total we have 5 latitudinal
and 11 longitudinal zones. Then we find out, e.g., that out of the 50
languages in the longitudinal zone between 160\degree{} East and 140\degree{}
West (covering many Pacific islands), 44 languages (88\%) have adjectives
after nouns.

Finally, we also created a new feature `latlon', which combines the
horizontal and vertical zone in one value. For instance, the Caucasian
languages (but not only them) would fall in latlon = 35--60;35--70.

\subsubsection{Development and Test Data}

As we have previously noted, approximately one half of the features in our
development data are visible, while the other half is masked; the same holds
for the test data. The merit of the visible part of the development data is
twofold. The obvious point is that known features $s_i$ of language $L$ serve
as input for prediction of a masked feature $t$ for language $L$. However, a
co-occurrence of known features $s_i$ and $s_j$ in one language may also
further improve our knowledge about feature correlations, which can be used
for predictions in other languages. Therefore we merged the known part of the
development data with the training data before creating the model and
applying it to the masked features.

During the evaluation phase of the shared task, we further enriched this
merged training data with the visible part of the blind test data.

\subsection{The Neural System}
\label{sec:neural}

\subsubsection{Architecture}
\begin{figure*}
    \centering
    \includegraphics{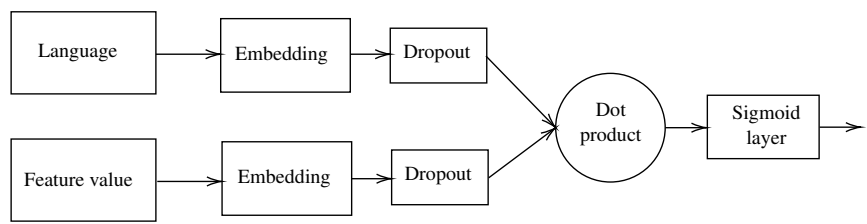}
    \caption{Neural system architecture}
    \label{fig:doc2vec}
\end{figure*}

In the neural system we use an architecture similar to Paragraph vectors
\cite{doc2vec}. The paragraph id is replaced by language id, and the words by
feature values. The overall architecture can be seen in \Fref{fig:doc2vec}.

\subsubsection{Preprocessing}
As feature input, we use all available features except the language code and
the country codes. We also cluster the geographical coordinates; but instead
of manually drawn zones as used in the Probabilistic System, we cluster the
points via k-Means and use the cluster id as a feature. The effect of number
of clusters and other hyper-parameters is discussed in the next section.

\subsubsection{Training}
The input to the neural network consists of a language id (which is a unique
identifier for each language) and a feature value (e.g. if feature is 81A
\textit{Order of Subject, Object and Verb} and its value is \textit{1~SOV},
then the feature value input to the neural network will be ``81A: Order of
Subject, Object and Verb: 1~SOV''). The goal of the neural network is a
binary prediction whether the given feature value belongs to the given
language. While training we select with 50\% probability a feature value that
belongs to the language, and with 50\% probability a feature value that does
not.

We use the Adam optimizer \cite{adam} with learning rate of $0.001$ for 200
epochs. We also run grid search on hyper-parameters to find the number of
clusters (1, 10, 50, 100, 150, 300, 500, 1000), embedding size (128, 512,
1024) and dropout rate (0, 0.3, 0.5). We select the best model based on
development data. We found out that the embedding size 512 with dropout rate
0.5 and 300 clusters yield the best accuracy of 73.9\% on development data.

\subsubsection{Prediction}
The prediction of an unknown feature value for a given language is done via
passing all possible values for the given feature to the network. We then
select the feature value with the highest output probability.

\subsubsection{Learned Embeddings}
To investigate whether the learned language embeddings encode meaningful
information, we run kNN ($k$ nearest neighbors) with cosine distance on the
embeddings to predict unknown features. With this approach we are able to
achieve accuracy of 68.10\% with majority vote on 33 nearest neighbors; this
is much higher than a kNN with Hamming distance (see \Sref{sec:other}).

The visualization of the learned language embeddings projected to 2D via TSNE
\cite{tsne} can be seen in appendix in \Fref{fig:families}. As we can see
from the visualization, there are some clear clusters, however, further work
would be needed to explain the clusters.

\subsubsection{Development and Test Data}
As in the Probabilistic system, we mix the visible part of the development
data (and later test data) with the training data. This is needed in order to
compute the language embeddings of the new languages.

\subsection{The Combined System}
\label{sec:combination}

We now describe the combination of the Probabilistic system and the Neural
system which we submitted for evaluation within the shared task.

\subsubsection{Analysis on Development Data}

As will be shown in Table~\ref{tab:deveval}, both the Probabilistic system
and the Neural system reach accuracies around 75\% on the development data,
with the Neural system being slightly better. However, a more detailed
analysis showed that while the accuracies are similar, the errors the systems
make are somewhat different: for 14\% of the masked feature values in the
development data, one of the systems predicts the correct value while the
other system does not (in absolute numbers there are 225 such cases in the
development data). An ensemble of the two systems could thus theoretically
reach an accuracy of up to 82\% if it were able to always select the correct
prediction. This motivates our efforts in finding a way to combine the
predictions of the two systems.

For this purpose, we utilize confidence scores produced by the two partial
systems. For the Probabilistic system, its $score_2$
(\Sref{sec:probabilistic:model}) serves as the confidence score for each
prediction; for the Neural system, we use the feature value probability
produced by the last layer of the neural network.

We analyzed the confidence scores on the 225 predictions on development data
where only one of the two systems predicted the correct feature value; in
48\% of cases it was the Probabilistic system, in 52\% it was the Neural
system.

Unfortunately, we found neither of the confidence scores to be a very good
predictor of the correctness of the prediction, with the Neural system
confidence score being somewhat more reliable. Specifically, low confidence
scores tend to indicate a potentially incorrect prediction for each of the
systems: on the least confident quarter of the predictions (25\% of the
analyzed predictions with lowest confidence score), the accuracy was only
around 40\% for each of the systems. High confidence scores are only
indicative for the Neural system (61\% accuracy on the most confident
quarter), for the Probabilistic system they seem to behave rather randomly
(accuracy on the most confident quarter is only 44\%.)

\subsubsection{Submitted Solution}

As the Neural system seems to perform better than the Probabilistic system,
our final solution is to return the prediction of the Neural system as the
final output, unless the Neural system confidence is very low and the
Probabilistic system confidence is not too low. Specifically, if the Neural
system confidence is below a threshold $T_N$ and the Probabilistic system
confidence is above a threshold $T_P$, we return the prediction from the
Probabilistic system, otherwise we return the prediction from the Neural
system.

Based on the development data analysis, we found the optimal threshold values
to be $T_N = 0.65$ and $T_P = 0.5$. On the development data, this leads to
returning the output of the Probabilistic system instead of the Neural system
in 6\% of cases in which the predictions of the two systems differ, leading
to an absolute gain of +1\% in accuracy. On the test data, the output of the
Probabilistic system was used in 10\% of the differing cases.

\subsection{Other Attempts}
\label{sec:other}

This section briefly describes some other approaches that we tried and
abandoned because their accuracy was too low.

\subsubsection{kNN}
We used a simple kNN with Hamming distance to find the nearest neighbors of a
language. The Hamming distance was calculated for each pair of languages as
the number of features whose values in the two languages do not match. If one
of the values for a given feature was unknown, we counted it as a mismatch.
We then found $k$ nearest neighbors of each language and filled the required
values using majority vote among the neighbors. This setup was able to
achieve accuracy of 62.28\% on development set using 22 nearest neighbors. We
have also tried different fallbacks such as most common value of a feature
overall, most common value in a genus or most common value in a family in
case of not finding any value among the neighbors. However, we found out that
these fallbacks had little to none impact on the final accuracy of the model.

\subsubsection{Feed-forward Neural Network}
We have also tried the classical feed-forward network to predict the required
features. As input, we randomly masked some of the non-empty feature values.
We then embedded each feature value into the embedding space and concatenated
these feature embeddings into a single vector. After the concatenation layer,
we added a few feed-forward layers and the classification head for each
feature. We have tried various architectures and percentages of masked values
with no success. We were able to achieve an accuracy of 56.45\% on the
development set, which is worse than the accuracy of kNN.

\section{Evaluation}
\label{sec:eval}

The evaluation metric is a simple accuracy: the number of correctly predicted
feature values divided by the number of masked features. Failure to predict a
value of a masked feature would count as an incorrect prediction.
Nevertheless, our systems always predict all masked features, even if there
are few reliable clues and the system's confidence is low.

The accuracies of different setups are given in \Tref{tab:deveval}. The
accuracy on the test data is the official number computed by the shared task
organizers (Baseline and Combined) or by us using the official evaluation
script (Probabilistic and Neural partial results); all the other results are
computed on the development data. As a baseline, we evaluate a model where
each feature receives its overall most probable value, regardless of
language. Interestingly, the Probabilistic system outperforms the Neural
system on the test data (contrary to what we observed on the development
data) but the difference is again in the order of just a few wrong
predictions.


\begin{table}[t]
  \begin{center}
    \begin{tabular}{lcc}
      \textbf{System} & \textbf{Dev} & \textbf{Test} \\\midrule
      Baseline        & 53.45 & 51.39 \\ 
      Probabilistic   & 73.81 & 71.08 \\
      Neural          & 74.49 & 69.80 \\
      Combined        & 75.50 & \textbf{70.75} \\\midrule
      Feed-forward    & 56.45 \\
      kNN-Hamming     & 62.28 \\
      kNN-LangEmbed   & 68.10 \\
    \end{tabular}
    \caption{\label{tab:deveval}Accuracy of various models on the development and test data.}
  \end{center}
\end{table}

Accuracies computed on individual languages are skewed because for some
languages the system had to predict only one feature. If we look at
development languages where 10 or more features were masked, our
Probabilistic system achieved 100\% on four of them (Ukrainian, Uyghur, Wan
and Venda); at the other end of the scale, accuracy on the South American
language Uru is only 30\%. The system never failed on 34 features; the most
frequent of them is unsurprisingly (see \Sref{sec:probabilistic})
\textit{Minor morphological means of signaling negation}, followed by
\textit{Order of Relative Clause and Noun} and \textit{Postnominal relative
clauses}. Note however that these observations do not necessarily indicate
that a feature is easy or difficult to predict. The success depends heavily
on how frequently the feature co-occurs with a suitable correlated feature in
WALS, and whether our random masking selected the feature in presence or
absence of the correlated feature.


%

\section{Conclusion}

We have described three systems (two individual systems and one combined
system) that use the known features of a language to predict its remaining
features. The first system is based on conditional probability of feature
values, log frequency and mutual information. It achieves an accuracy of
73.81\% on the development data. The second system uses a neural network to
train language and feature embeddings, and to predict a feature value's
probability based on the embeddings. It achieves slightly better results than
the probabilistic model on the development data, but it is slightly worse on
the test data. Finally, we use the scores produced by the two systems
together with their predictions to decide which of the systems knows the
correct answer in which situation. This combined system achieves 75.50\% on
the development data and 70.75\% on the test data.

Our code is available at \url{https://github.com/ufal/ST2020}.

\section*{Acknowledgments}

This work was supported
by the LINDAT/ CLARIAH-CZ project of the Ministry of Education, Youth and
Sports of the Czech Republic (project LM2018101).

\bibliography{anthology,emnlp2020}

\begin{thebibliography}{8}
\expandafter\ifx\csname natexlab\endcsname\relax\def\natexlab#1{#1}\fi

\bibitem[{Bjerva et~al.(2020)Bjerva, Salesky, Mielke, Chaudhary, Celano, Ponti,
  Vylomova, Cotterell, and Augenstein}]{st-overview2020sigtyp}
Johannes Bjerva, Elizabeth Salesky, Sabrina Mielke, Aditi Chaudhary, Giuseppe
  G.~A. Celano, Edoardo~M. Ponti, Ekaterina Vylomova, Ryan Cotterell, and
  Isabelle Augenstein. 2020.
\newblock {SIGTYP 2020 Shared Task: Prediction of Typological Features}.
\newblock In \emph{Proceedings of the Second Workshop on Computational Research
  in Linguistic Typology}. Association for Computational Linguistics.

\bibitem[{Daum{\'e}~III and Campbell(2007)}]{daume2007}
Hal Daum{\'e}~III and Lyle Campbell. 2007.
\newblock \href {https://www.aclweb.org/anthology/P07-1009/} {A {Bayesian}
  model for discovering typological implications}.
\newblock In \emph{Proceedings of the 45th Annual Meeting of the Association
  for Computational Linguistics}, pages 65--72, Praha, Czechia. Association for
  Computational Linguistics.

\bibitem[{Dryer and Haspelmath(2013)}]{wals}
Matthew~S Dryer and Martin Haspelmath. 2013.
\newblock \href {http://wals.info/} {The {World Atlas of Language Structures}
  online}.

\bibitem[{Greenberg(1963)}]{greenberg}
Joseph~H. Greenberg. 1963.
\newblock Some universals of grammar with particular reference to the order of
  meaningful elements.
\newblock In Joseph~H. Greenberg, editor, \emph{Universals of Language}, pages
  110--113. MIT Press, London.

\bibitem[{Kingma and Ba(2014)}]{adam}
Diederik~P. Kingma and Jimmy Ba. 2014.
\newblock \href {http://arxiv.org/abs/1412.6980} {Adam: A method for stochastic
  optimization}.
\newblock Cite arxiv:1412.6980Comment: Published as a conference paper at the
  3rd International Conference for Learning Representations, San Diego, 2015.

\bibitem[{Le and Mikolov(2014)}]{doc2vec}
Quoc~V. Le and Tomas Mikolov. 2014.
\newblock \href {http://arxiv.org/abs/1405.4053} {Distributed representations
  of sentences and documents}.
\newblock \emph{CoRR}, abs/1405.4053.

\bibitem[{van~der Maaten and Hinton(2008)}]{tsne}
Laurens van~der Maaten and Geoffrey Hinton. 2008.
\newblock \href {http://www.jmlr.org/papers/v9/vandermaaten08a.html}
  {Visualizing data using {t-SNE}}.
\newblock \emph{Journal of Machine Learning Research}, 9:2579--2605.

\bibitem[{O{'}Horan et~al.(2016)O{'}Horan, Berzak, Vuli{\'c}, Reichart, and
  Korhonen}]{ohoran2016}
Helen O{'}Horan, Yevgeni Berzak, Ivan Vuli{\'c}, Roi Reichart, and Anna
  Korhonen. 2016.
\newblock \href {https://www.aclweb.org/anthology/C16-1123} {Survey on the use
  of typological information in natural language processing}.
\newblock In \emph{Proceedings of {COLING} 2016, the 26th International
  Conference on Computational Linguistics: Technical Papers}, pages 1297--1308,
  Osaka, Japan. The COLING 2016 Organizing Committee.

\end{thebibliography}
\bibliographystyle{acl_natbib}

\appendix

\section{Appendix}
\label{sec:appendix}

\begin{landscape}
\begin{figure}
    \centering
    \includegraphics[width=22cm]{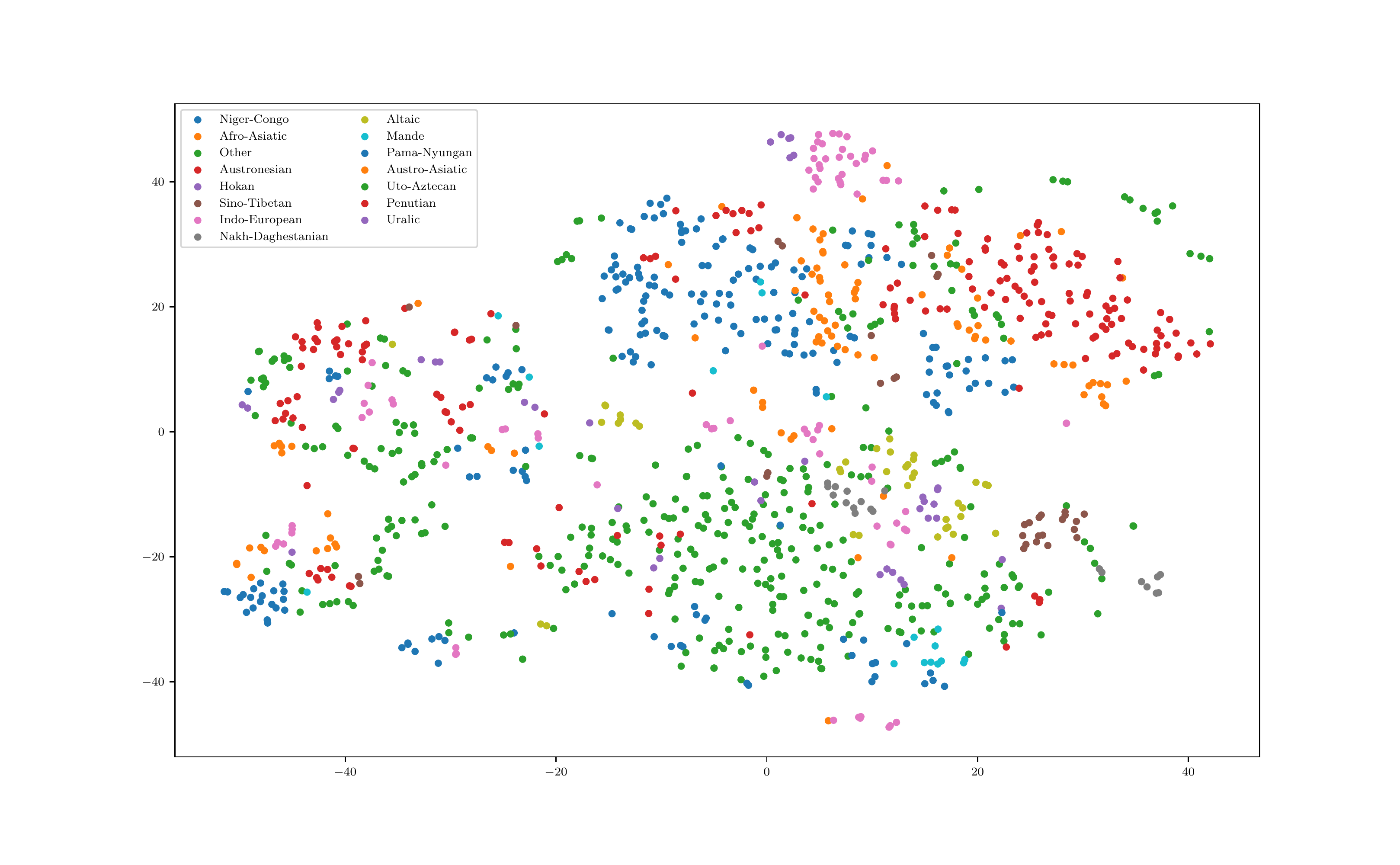}
    \caption{Visualisation of learned embeddings by family.}
    \label{fig:families}
\end{figure}
\end{landscape}

\end{document}